# A REASON MAINTENANCE SYSTEM DEALING WITH VAGUE DATA

B.Fringuelli, S.Marcugini, A.Milani, S.Rivoira
Dipartimento di Matematica
Università di Perugia
via Vanvitelli, 1
06100 Perugia, ITALY

## Abstract

A reason maintenance system which extends an ATMS through Mukaidono's fuzzy logic is described. It supports a problem solver in situations affected by incomplete information and vague data, by allowing nonmonotonic inferences and the revision of previous conclusions when contradictions are detected.

## INTRODUCTION

Any reasoning system must deal with belief revision at some extent.

In recent years truth maintenance systems have been proposed as powerful tools able to perform belief revision at a general level.

These systems can be viewed as constraint propagation mechanisms which tell a problem solver what things it is currently obliged to believe, given a single set of premises and a set of deduction constraints, some of which may be nonmonotonic.

Justification-based TMS (Doyle 1979) (McAllester 1980) (McDermott 1983) maintain a single context of belief and support nonmonotonic justifications, while assumption-based TMS (de Kleer 1986) (Martins 1988) avoid the restriction that the overall set of premises is contradiction free, maintaining multiple contexts of belief.

Early truth maintenance systems dealt with certain beliefs only, but several successive works extended them in order to allow handling of some kind of uncertainty.

De Kleer and Williams (de Kleer 1987) have assigned probabilities to assumptions in an ATMS which diagnoses multiple mutually indipendent faults.

Falkenheiner (Falkenheiner 1988) has introduced Dempster-Shafer theory into Doyle's TMS.

D'Ambrosio (D'Ambrosio 1989) has used an ATMS to compute beliefs for a special case of the Dempster-Shafer model.

Provan (Provan 1989) has incorporated belief functions into ATMS and Laskey and Lehner (Laskey 1989) have shown that any Dempster-Shafer inference system can be represented in a ATMS by attaching probabilities to assumptions that represent hypotheses in a background frame.

Dubois et al. (Dubois 1990) extended an ATMS in order to handle uncertainty, pervading justifications or grading assumptions, represented in the framework of possibility and necessity measures.

In this paper we describe a Fuzzy Truth Maintenance System (FTMS) obtained by extending an ATMS through fuzzy logic.

The general idea and motivations of our approach are very close to those of Dubois et al. (Dubois 1990).

The main difference from their work lies in the fact that in our system propositions involve vague predicates which may have intermediary degrees of truth and the underlying logic is truth-functional, while Dubois et al. consider propositions which are true or false, but due to the lack of precision of the available information it can only be estimated to what extent it is possible or necessary that a proposition is true.

In the following the basic definitions and properties of the adopted fuzzy logic are reported and, successively, definitions and functionalities of the FTMS are discussed.

## MANY-VALUED LOGICS AND RESOLUTION

The work described in this paper is part of a research aiming to compare existing theories of uncertainty (both logical and probabilistic) from the viewpoint of the efficiency of inference rules and revision mechanisms.

Since the resolution principle (Robinson 1965) encompasses several inference rules in classical logic (modus ponens, modus tollens, disjunctive and hypothetical syllogisms, constructive and desctructive dilemmas) and it is widely used in reasoning systems, we firstly focused our attention on extensions of the resolution principle dealing with some kind of uncertainty.

Dubois and Prade (Dubois 1987) extended the resolution principle in the case of uncertain propositions where the uncertainty involves non-vague predicates and it is modeled in terms of necessity measures.

The first attempt to a theory of fuzzy resolution was proposed by Lee (Lee 1972) for a fuzzy logic defined as follows.

Let [S] denote the truth value of a formula S:

$[S] \in [0,1]$

$[\neg S] = 1 - [S]$

$[R \vee S] = max([R],[S])$

$[R \wedge S] = min([R],[S])$

$[R \rightarrow S] = [\neg R \vee S] = max(1-[R],[S]).$

An interpretation I is said to satisfy a formula S if $[S] \geq 0.5$ under I.

The resolution principle corresponds to the following rule of inference:

let    $S_1 = x \vee L_1$;

$S_2 = \neg x \vee L_2$;

a logical consequence of $S_1 \wedge S_2$ is the resolvent:

$R(S_1, S_2) = L_1 \vee L_2$

and, in fuzzy logic, if $[S_1 \wedge S_2] > 0.5$ then:

$0.5 < [S_1 \wedge S_2] \leq [R(S_1, S_2)] \leq [S_1 \vee S_2].$

Basically Lee proved that the resolution principle is complete in fuzzy logic and if every clause in a set has a truth-value greater than 0.5, then all the logical consequences obtained by repeatedly applying the resolution principle will have truth-value at least equal to the most unreliable clause, but never exceeding the truth value of the most reliable one.

These results were extended to a more general case by Mukaidono (Mukaidono 1982) (Mukaidono 1989) which allowed the truth value of all the clauses to be taken in the closed interval [0,1], introducing an inference strategy for fuzzy Prolog based on the following definitions.

The confidence c(S) of a formula S is defined as

$c(S) = ([S] - 0.5) * 2$

and the fuzzy resolution principle asserts that the *confidence of resolution* $c_r$ of the resolvent $R(S_1, S_2)$ is:

$c_r(R(S_1, S_2)) = (max([x],[\neg x]) - 0.5) * 2 = |c(x)|$

where x is the key predicate in the resolution.

If $S_2 = R(S3, S4)$ then

$c_r(R(S1, S2)) = min(c_r(S2), |c(x)|).$

The definition $[R \rightarrow S] = [\neg R \vee S] = max(1-[R],[S])$ adopted for implication in fuzzy logic allows the inference of S from R ( or $\neg R$ from $\neg S$) only when

$[R \rightarrow S] \geq [\neg R]$ (or $[R \rightarrow S] \geq [S]$ respectively).

This resolution principle is proved to be complete and significant for any truth value in the closed interval [0,1].

The confidence of resolution of an inferred formula S represents the degree of derivability of S from the formulas used in the inference process.

Mukaidono introduced an additional concept for implication (weight of rule) defined as the product of the confidence values of premise and conclusion:

$w_{R \rightarrow S} = c(R) * c(S)$

The weight of rule (usually defined as a closed interval) represents the degree of truth of an implication and it establishes the applicability of the rule, given the confidence of either the premise or the conclusion.

In fact it is easy to prove that a rule $R^w \text{->} S$ can be applied if and only if:

$|w| \leq |c(R)|$  and  $|w| \leq |c(S)|$.

According to the previous definitions it is possible to derive from a given set of fuzzy Horn clauses all the fuzzy logical consequences, together with their confidences of resolution.

The following example shows the inference mechanism applied to propositional clauses (first order predicate logic can be easily obtained by introducing unification ).

From:

r1)  $A{\rightarrow}B$    $\{w1 = 0.3\}$

r2)  $B{\rightarrow}C$    $\{w2 = -0.4\}$

r3)  $A{\rightarrow}D$    $\{w3 = -0.7\}$

r4)  $D{\rightarrow}C$    $\{w4 = 0.1\}$

r5)  $A$        $\{[A] = 0.8\}$

it is possible to derive:

$c(A) = ([A]-0.5) * 2 = 0.6;$    $c_r(A) = 1;$        (from r5)

$c(B) = w1/c(A) = 0.5$     $c_r(B) = \min(c_r(A),|c(A)|) =$ 0.6   (from r1, r5)

Since  $|w3| > c|(A)|$, r3 cannot be applied.

$c(C) = w2/c(B) = -0.8; c_r(C) = \min (c_r(B), |c(B)|) =$ 0.5  (from r5,r1,r2)

The fuzzy proposition  C  is therefore a logical consequence of proposition S r5, r1 and r2. The inferred truth-value of C is:

$$[C] = c(C)/2 + 0.5 = 0.1$$

while its confidence of resolution, that represents the degree of derivability of C from the axioms, is:

$c_r(C) = 0.5.$

The confidence c(P) of a conclusion P and its confidence of resolution $c_r(P)$ can be combined to give the confidence of resolved consequence:

$crc(P) = c(P) * c_r(P).$

# DEFINITION OF A FUZZY TRUTH MAINTENANCE SYSTEM

Extending De Kleer 's definition of ATMS (de Kleer 1986), we define an FTMS in the following way.

Every fuzzy formula introduced or derived by the attached problem solver corresponds to an FTMS **node**.

A special kind of node is represented by the atom $\perp$, corresponding to "falsity", for which $[\perp] = 0$ holds in any interpretation.

A **justification** is a triple:

$$<j,c(n),c_r(n)>$$

where  j: $x_1,x_2,...,x_m$ -> n is a propositional Horn clause asserting that the consequent node n is derivable from the conjunction of the antecedent nodes $x_1,...,x_m$ and c(n) and $c_r(n)$ are respectively the confidence and the confidence of resolution established by j for the node n.

A justification $<j,-1,c_r(\perp)>$, where the derived node is falsity, is communicated by the problem solver every time a contradiction is detected.

An **assumption** is a self-justifying node representing the decision of introducing an hypothesis; it is connected to the assumed data through justifications.

An **environment** is a set of logically conjuncted assumptions .

An environment E has **consistency** cs(E) equal to the opposite of the maximal confidence of resolved consequence with which falsity can be derived from E and the current set J of justifications:

$$cs(E) = - \max_{j} c_r(\perp)_E$$

An FTMS **context** is defined as the set formed by the assumptions of an environment and all the nodes derivable from those assumptions.

The goal of FTMS is to efficiently update the contexts when new assumptions or justifications are provided by the problem solver.

This goal is achieved by associating with every node a description (**label**) of every context in which the node holds.

More formally, a label $L_n$ of the node n is defined as the set of all the environments from which n can be derived:

$$L_n = \{E_i : E_i \Rightarrow n\}$$
$$\quad\quad\quad\quad j$$

In order to save space and time, a problem solver may wish to consider only environments whose consistency is greater than some threshold $\alpha$ and/or from which nodes can be derived with a degree of derivability greater than some threshold $\beta$ , where $\alpha$ and $\beta$ depend on the problem domain.

Therefore, given the two lower bounds $\alpha$ and $\beta$ four important properties can be defined for the labels:

a label $L_n$ is $\alpha$-**consistent** if the consistency of each of its environments is not less than $\alpha$;

a label $L_n$ is $\beta$-**sound** if n is derivable from each of its environments with a confidence of resolution not less than $\beta$;

a label $L_n$ is $\alpha$-$\beta$-**complete** if every $\alpha$-consistent environment from which n can be derived with a confidence of resolution not less than $\beta$ is a superset of some environment in $L_n$;

a label $L_n$ is **minimal** if no environment $E_i$ in $L_n$ is a superset of another environment $E_k$ in $L_n$ with $crc_i(n) \leq crc_k(n)$.

The task of FTMS is to ensure that each label in each node is $\alpha$-consistent, $\beta$-sound, $\alpha$-$\beta$-complete and minimal with respect to the current set of justifications .

This task is performed by invoking the following label-updating algorithm every time the problem solver adds a new justification.

Firstly the justification is recorded and then the new label and new confidence values are evaluated for the justified node.

If the new label or confidence values are different from the old ones, the algorithm considers the datum associated with the node. If it is not the falsity ,then the updating process recursively involves the labels and confidences of all the consequent nodes .

If the newly justified node is falsity ,the consistency of each environment in the label is computed and the environment database is updated.

It is worth noticing that the revision of node confidences can make no more significant previously applied rules , forcing the system to retract the corresponding justifications.

Justifications are made retractable by conjoining them with extra assumptions which represent their defeasability.

Only the minimal environment database (MEDB) is maintained in the sense that an environment $E_2$ is recorded in the database only if no environment $E_1$ exists such that:

$$(E_1 \subset E_2) \text{ and } (cs(E_1) > cs(E_2)).$$

In contrast with ATMS, where inconsistent environments are removed from every node label, FTMS always keeps the environments in their labels, since consistency can be changed by successive justifications.

FTMS maintains for each fuzzy formula S introduced or derived by an attached problem solver the following information:

-the truth value of S, represented by the confidence established by the justifications of the corresponding node;

-the degree of derivability of S from the current knowledge, represented by the confidence of resolution of the corresponding node;

-the minimal set of environments from which S can be derived, together with their consistency values.

At each step of the reasoning process, the problem solver can therefore rank the partial solutions currently available on the basis of several ordering criteria (truth value, degree of derivability, consistency of the hypotheses), discarding or eliminating solutions which are not enough founded.

The main mechanisms for updating labels and confidences and their possible effects on the reasoning process are illustrated by the following example.

Let us suppose that the problem solver, on the basis of its own domain knowledge and inference procedures, has already derived and communicated to FTMS the justifications reported in figure 1 (where $\pi$, $\rho$, $\sigma$, $\tau$ are assumptions and $\perp$ indicates falsity) from which FTMS has determined the labels and the minimal environment database reported in figure 2.

The consequent net of dependencies between assumptions and derived propositions is shown in figure 3.

$R_5$:  A,B → E, {w5= 0.2}

$R_6$:  A,C → F, {w6= 0.3}

$R_7$:  E,F → H, {w7= 0.4}

$R_8$:  C,D → F, {w8= 0.4}

R9:  B,F → E, {w9= 0.3}

$R_{10}$:  C,B → ⊥, {w10= 0.2}

$R_{11}$:  F,D → G, {w11= 0.2}

$R_{12}$:  A,H → G, {w12= 0.4}

$R_{13}$:  F,G → E, {w13= 0.4}

$R_{14}$:  D,E → ⊥, {w14= 0.5}

figure 1a: set of inference rules

$J_1$:  < π → A, c(A) = 0.6,  $c_r$(A) = 1>

$J_2$:  < ρ → B, c(B) = 0.4,  $c_r$(B) = 1>

$J_3$:  < σ → C, c(C) = 0.4,  $c_r$(C) = 1>

$J_4$:  < τ → D, c(D) = 0.4,  $c_r$(D) = 1>

$J_5$:  <A,B → E, c(E) = 0.5,  $c_r$(E) = 0.5>

$J_6$:  <A,C → F, c(F) = 0.75,  $c_r$(F) = 0.75>

$J_7$:  <E,F → H, c(H) = 0.8,  $c_r$(H) = 0.8>

$J_8$:  <C,D → F, c(F) = 1,    $c_r$(F) = 1>

$J_9$:  <B,F → E, c(E) = 0.75,  $c_r$(E) = 0.75>

$J_{10}$:  <C,B → ⊥, c(⊥) = 0.5,  $c_r$(⊥) = 0.5>

$J_{11}$:  <F,D → G, c(G) = 0.75,  $c_r$(G) = 0.75>

$J_{12}$:  <A,H → G, c(G) = 0.76, $c_r$(G) = 0.53>

figure 1b: a current set of justifications

$L_A$ = {[(π), cs=1]}

$L_B$ = {[(ρ), cs=1]}

$L_C$ = {[(σ), cs=1]}

$L_D$ = {[(τ), cs=1]}

$L_E$ = {[(π,ρ), cs=1], [(π,σ,τ), cs=-0.5]}

$L_F$ = {[(π,σ), cs=1], [(σ,τ), cs=1]}

$L_G$ = {[(π,ρ,σ), cs=-0.5], [(σ,τ), cs=1]}

$L_H$ = {[(ρ,σ,τ), cs=-0.5]}

figure 2a: the label of each node

MEDB: [(ρ,σ), cs=-0.5]

figure 2b: the minimal environment database

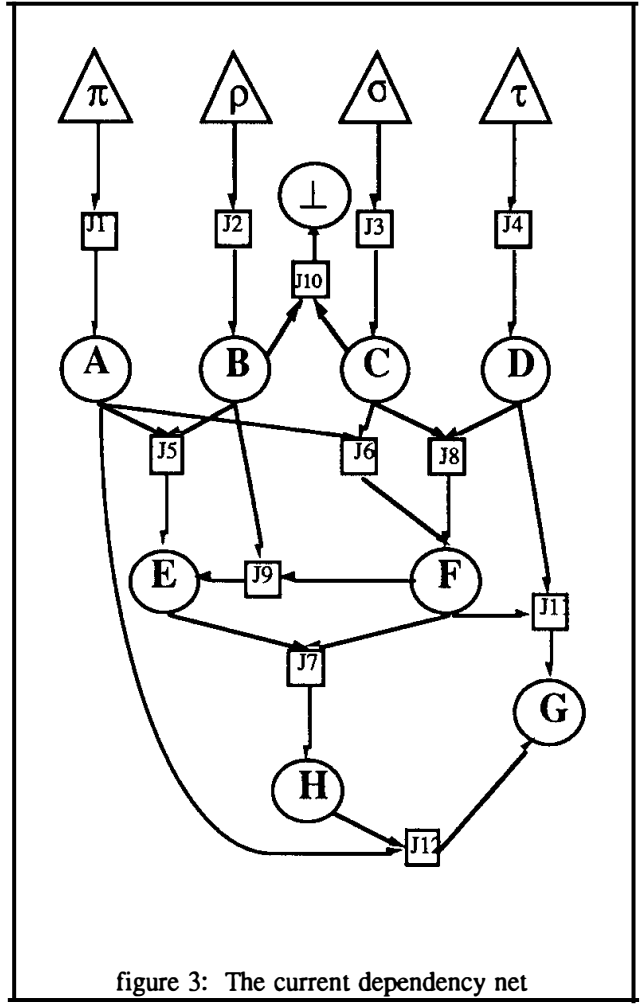

figure 3:  The current dependency net

Let now the problem solver adds the justification (see figure 4):

$J_{13}$  <F,G → E, c(E) = 0.8, $c_r$(E) = 0.75>

Since a new confidence value for E is introduced, it is necessary to update the truth-values of all the consequent nodes. In this case the updating process terminates after the new values for H have been evaluated, because the connfidence in G is not affected.

c(H) = 0.5    $c_r$(H) = 0.5

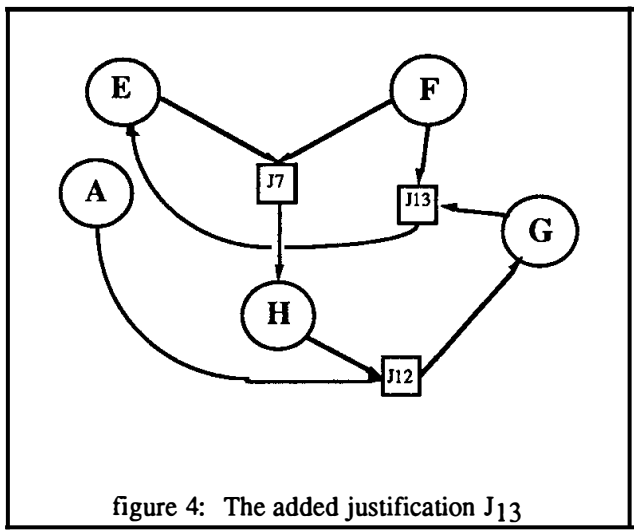

figure 4:  The added justification $J_{13}$

The effect of $J_{13}$ on the labels is the following:

$L_E = \{[(\pi,\rho), cs=1], [(\sigma,\tau), cs=1]\}$

$L_H = \{[(\pi,\rho,\sigma), cs=-0.5], [(\sigma,\tau), cs=1]\}$

$L_G = \{[(\pi,\rho,\sigma), cs=-0.5], [(\sigma,\tau), cs=1]\}$

Let us finally suppose that a new contradiction, represented by the justification $J_{14}$, is detected by the problem solver:

$J_{14}:$   $<D,E \rightarrow \perp, c(\perp) = 0.4, c_r(\perp) = 0.4>$

This justification modifies the minimal environment database, introducing two new entries:

MEDB:

$[(\sigma,\tau), cs=-0.5]$

$[(\pi,\rho,\tau), cs=-0.4]$

$[(\sigma,\tau), cs=-0.4]$

Therefore the new labels become:

$L_E = \{[(\pi,\rho), cs=1], [(\sigma,\tau), cs=-0.4]\}$

$L_F = \{[(\pi,\sigma), cs=1], [(\sigma,\tau), cs=-0.4]\}$

$L_G = \{[(\pi,\rho,\sigma), cs=-0.5], [(\sigma,\tau), cs=-0.4]\}$

$L_H = \{[(\pi,\rho,\sigma), cs=-0.5], [(\sigma,\tau), cs = -0.4]\}$

## CONCLUSION

The system described in this paper supports a problem solver in the task of selecting among several alternatives in situations affected by incomplete information, uncertain knowledge and vague data.

FTMS allows the problem solver to make nonmonotonic inferences, revising previous conclusions if contradictions are detected.

Every derived belief is associated with three parameters: a confidence which shows how much it is true, a confidence of resolution, which tells to what extent it is derivable from the current knowledge, and a consistency, which represents the degree of contradiction of the hypotheses which it relies on.

Dependencies between beliefs are recorded so that when new information is supplied, only the affected beliefs are involved in the updating process.

FTMS has been successfully implemented in Prolog.

## Acknowledgements.

The authors wish to thank Settimo Termini fro the valuable discussions about many valued logics.

This work has been supported by P.F. Robotica - Consiglio Nazionale delle Ricerche grant n.90.00556.67 and M.U.R.S.T.- 40% "Tecniche di Ragionamento Automatico e Sistemi Intelligenti".

## References

B.D'Ambrosio (1989). A Hybrid Approach to Reasoning Under Uncertainty. In L.N.Kanal, T.S.Levitt, J.F.Lemmer (eds.) *Uncertainty in Artificial Intelligence: 3rd Conference*, North-Holland, pp.267-283.

J.Doyle (1979). A Truth Maintenance System. In *Artificial Intelligence*, **12** (3) ,pp 231- 272.

D.Dubois, J.Lang, H.Prade (1990). Handling Uncertain Knowledge in an ATMS Using Possibilistic Logic. In *Proceeding of ECAI Workshop on Truth Maintenance Systems* ,Stockolm.

D.Dubois, H.Prade (1987). Necessity Measures and the Resolution Principle. In *IEEE Trans. on Systems, Man and Cyberneticss* , vol.SMC-17, n.3, pp.474-478.

B.Falkenheiner (1988). Towards a General Purpose Belief Maintenance System. In J.F.Lemmer, L.N.Kanal

(eds.) *Uncertainty in Artificial Intelligence: 2nd Conference*, North-Holland, pp.125-132.

J.de Kleer (1986). An Assumption-based TMS. In *Artificial Intelligence* , 28 (2) , pp.127-162.

J.de Kleer, B.C.Williams (1987). Diagnosing Multiple Faults. In *Artificial Intelligence* , 32, pp. 97-130.

K.B.Laskey, P.E. Lehner(1989). Assumptions, Beliefs and Probabilities. In *Artificial Intelligence* 41, pp. 65-77.

R.C.T.Lee (1972). Fuzzy  Logic and the Resolution Principle. In *Journal of ACM*, 19 (1), pp.109-119.

J.P.Martins, S.C.Shapiro (1988). A Model for Belief Revision. In *Artificial Intelligence* , 35, pp. 25-79.

D.McAllester (1980). An Outlook on Truth Maintenance. In *AI Memo* 551, AI Lab., MIT, Cambridge (MA).

D.McDermott (1983). Context and Data Dependencies. In *A Synthesis*, IEEE Trans. Pattern  Anal.Mach. Intell., 5 (3), pp. 237-246.

M.Mukaidono (1982 ). Fuzzy Inference of Resolution Style. In R.R.Yager (Ed.) *Fuzzy Set and Possibility Theory*, Pergamon Press, New York , pp. 224-231.

M.Mukaidono, Z. Shen, L. Ding (1989 ). Fundamentals of Fuzzy Prolog. In *International Journal of Approximate Reasoning*, 3, pp. 179-193.

G.M. Provan (1989). An Analysis of ATMS-based Techniques for Computing Dempster-Shafer Belief Functions. In *Proceedings of the.9th IJCAI* , Detroit Aug. 1989 , pp. 1115-1120.

J.A.Robinson (1965). A Machine-oriented Logic Based on the Resolution Principle. In *Journal of ACM*, 12 (1), pp. 23-41.